\documentclass[journal]{IEEEtran}

\usepackage[export]{adjustbox}
\usepackage{algorithmic}
\usepackage{algorithm}
\usepackage{amsmath, amssymb, amsfonts, amsthm}
\usepackage{array}
\usepackage{booktabs}
\usepackage{colortbl, xcolor}
\usepackage{cite}
\usepackage{graphicx}
\usepackage{multirow}
\usepackage{nicematrix}
\usepackage{soul}
\usepackage[caption=false]{subfig}
\usepackage{tabularx}
\usepackage{textcomp}
\usepackage{tikz}
\usepackage{threeparttable}
\usepackage{wrapfig}
\usepackage{xcolor}
\definecolor{steelblue}{HTML}{4682B4}
\colorlet{baseline}{orange!20}
\colorlet{our}{cyan!15}

\usepackage{xkcdcolors}
\usepackage[pdfusetitle]{hyperref}
\hypersetup{
  colorlinks=true,
  linkcolor=xkcdDarkSkyBlue,
  urlcolor=xkcdDullPink,
  citecolor=xkcdDarkSkyBlue,
}

\DeclareMathSizes{10}{8}{8}{5}

\DeclareMathOperator{\sign}{sign}

\newtheorem{theorem}{Theorem}
\newtheorem{lemma}{Lemma}

\allowdisplaybreaks

\newcommand\copyrighttext{%
  \footnotesize \textcopyright 2026 IEEE. All rights reserved, including rights for text and data mining, and training of artificial intelligence and similar technologies. Personal use is permitted, but republication/redistribution requires IEEE permission. DOI: \href{https://doi.org/10.1109/TNNLS.2026.3718377}{10.1109/TNNLS.2026.3718377}.}
\newcommand\copyrightnotice{%
\begin{tikzpicture}[remember picture,overlay]
\node[anchor=south,yshift=10pt] at (current page.south) 
  {\fbox{\parbox{\dimexpr\textwidth-\fboxsep-\fboxrule\relax}{\copyrighttext}}};
\end{tikzpicture}%
}

\begin{document}

\title{Deep Residual Echo State Networks: exploring residual orthogonal connections in untrained Recurrent Neural Networks}

\author{Matteo Pinna, Andrea Ceni, Claudio Gallicchio%
\thanks{
    Published in IEEE Transactions on Neural Networks and Learning Systems.
    The authors are with the Department of Computer Science, University of Pisa, 56127 Pisa, Italy (e-mail: matteo.pinna@di.unipi.it; andrea.ceni@unipi.it; claudio.gallicchio@unipi.it).
    Code is available at \href{https://github.com/nennomp/deepresesn}{github.com/nennomp/deepresesn}.
}}

\maketitle
\copyrightnotice
\begin{abstract}
Echo State Networks (ESNs) are a particular type of untrained Recurrent Neural Networks (RNNs) within the Reservoir Computing (RC) framework, popular for their fast and efficient learning. However, traditional ESNs often struggle with long-term information processing. In this paper, we introduce a novel class of deep untrained RNNs based on temporal residual connections, called Deep Residual Echo State Networks (DeepResESNs). We show that leveraging a hierarchy of untrained residual recurrent layers significantly boosts memory capacity and long-term temporal modeling. For the temporal residual connections, we consider different orthogonal configurations, including randomly generated and fixed-structure, and study their effect on network dynamics. A thorough mathematical analysis outlines necessary and sufficient conditions to ensure stable dynamics within DeepResESN. Empirically, the proposed approach consistently outperforms traditional shallow and deep RC on a variety of time series tasks. Overall, DeepResESN offers a promising approach for designing hierarchical ESNs with better prediction accuracy on long sequences, without sacrificing the computational advantages that make RC attractive.
\end{abstract}


\section{Introduction}
\label{sec:introduction}
\IEEEPARstart{D}{eep} Neural Networks (DNNs) have driven major advances in fields such as computer vision and natural language processing, thanks to their capacity to learn hierarchical data representations through the composition of multiple non-linear layers \cite{lecun2015deep, simonyan2015deep, vaswani2017attention}. Recurrent Neural Networks (RNNs) similarly benefit from depth, both in their architectural design and through their temporal unfolding, which effectively creates deep computational graphs across time steps. This temporal depth makes RNNs powerful for modeling sequential data, with applications spanning areas such as control and robotics \cite{shi2021novel}, but also introduces challenges analogous to those encountered in deep feedforward networks, such as vanishing and exploding (V/E) gradients \cite{bengio1994gradient, glorot2010understanding}. These issues, compounded by increased computational demands, hinder the training of very deep recurrent models. Randomized untrained models offer an alternative to gradient-based training. In particular, the Reservoir Computing (RC) framework \cite{verstraeten2007reservoir} uses fixed recurrent dynamics and trains only a readout layer, reducing training complexity. Deep Echo State Networks (DeepESNs) \cite{gallicchio2017deepesn} extend this idea by stacking multiple untrained reservoirs to build temporal hierarchies. However, while RC models bypass backpropagation, they are still susceptible to signal degradation or amplification in the forward pass, both in the temporal and architectural dimension. These effects can impair the stability and expressiveness of the model, limiting its effectiveness on complex temporal tasks.
\begin{figure}[t]
    \includegraphics[scale=1.0, center]{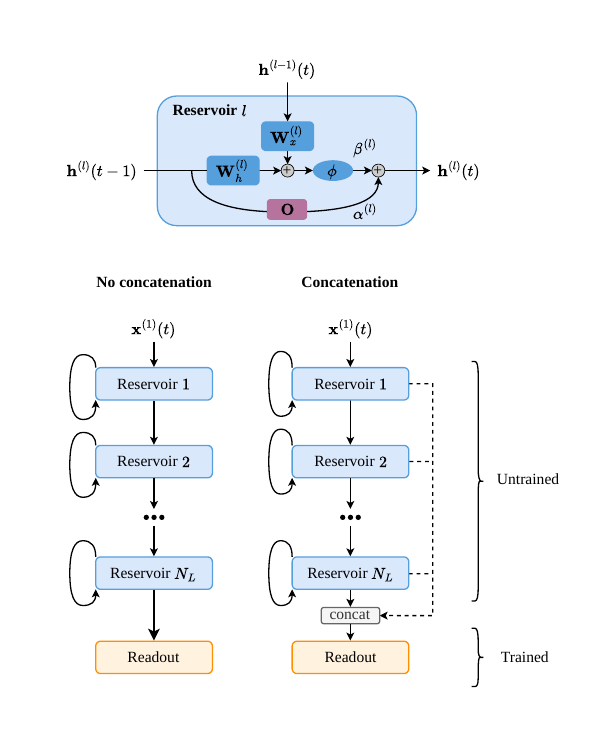}
    \caption{Architectural organization of the proposed DeepResESN. (\emph{top}) Structure of a generic $l$-th reservoir layer in a DeepResESN. The reservoir structure (shown in blue) consists of an input weight matrix $\mathbf{W}_x^{(l)}$, a recurrent weight matrix $\mathbf{W}_h^{(l)}$, and a non-linear activation function $\phi$. The temporal residual connection (shown in purple) is modulated by an orthogonal matrix $\mathbf{O}$. The temporal residual and non-linear paths are scaled by positive coefficients $\alpha^{(l)}$ and $\beta^{(l)}$, respectively. (\emph{bottom}) Complete illustration of a DeepResESN architecture with $N_{L}$ reservoir layers. The first layer acts as a residual reservoir in a traditional shallow architecture and is fed the external input $\mathbf{x}^{(1)}$. Subsequent layers receive as input the output of the previous reservoir, $\mathbf{h}^{(l-1)}$. The readout may be fed either the final layer states or the concatenation of states from all layers. See Section~\ref{sec:deepresesn} for details.}
    \label{fig:deepresesn_architecture}
\end{figure}
\IEEEpubidadjcol
To address analogous issues in feedforward architectures, residual connections have proven highly effective. Introduced in the context of convolutional networks, Residual Networks (ResNets) \cite{he2016resnet} enhance information flow by incorporating identity-based skip connections, thereby enabling the training of very deep models. Although originally proposed for fully-trainable networks, the core principle, facilitating signal propagation through additional paths, is broadly applicable and conceptually appealing for mitigating dynamical degradation in untrained settings as well. Although residual pathways are a promising solution, they remain relatively underexplored in recurrent architectures. Prior work has explored temporal residual connections in trainable RNNs \cite{chang2017dilated, ceni2025random} and shallow untrained RNNs \cite{ceni2024residual, pinna2025residual}. Their extension to deep untrained RNNs remains unexplored.

In this paper, we introduce Deep Residual Echo State Networks (DeepResESNs), a novel class of deep untrained RNNs that unify the hierarchical representation capabilities of DeepESNs with the enhanced temporal signal propagation of ResESNs, thereby providing a principled generalization of both architectures. The architecture is graphically highlighted in Fig.~\ref{fig:deepresesn_architecture}. Specifically, (i) we propose a deep RC model in which each untrained recurrent layer is augmented with a temporal residual connection, governed by configurable mappings such as random orthogonal, cyclic, or identity transformations; (ii) we explore the effect of each transformation on the network's dynamics, leveraging spectral frequency analysis tools; (iii) we provide a comprehensive mathematical analysis of DeepResESN dynamics, deriving necessary and sufficient conditions for stability and contractivity, and extending the Echo State Property (ESP) to the deep residual case; (iv) we evaluate DeepResESNs on time series tasks spanning memory, forecasting, and classification. The results demonstrate consistent improvements over both shallow and deep RC baselines, especially in settings requiring long-term temporal modeling.

The remainder of this paper is organized as follows. Section~\ref{sec:background} introduces background on RC. Section~\ref{sec:deepresesn} presents the DeepResESN architecture and the spectral frequency analysis. Section~\ref{sec:mathematics} presents our theoretical analysis. Section~\ref{sec:experiments} is dedicated to the experiments. Section~\ref{sec:conclusions} concludes the paper. Appendix~\ref{sec:proofs} is dedicated to mathematical proofs.

\section{Reservoir Computing}
\label{sec:background}
RC-based neural networks consist of a large, untrained recurrent layer, called the \emph{reservoir}, and a trainable linear readout layer. The reservoir is randomly initialized subject to the Echo State Property (ESP) \cite{yildiz2012re}, a stability condition useful to guide initialization in ESNs, and then left untrained. The readout is the only trainable component and processes the reservoir's output. The underlying intuition is to exploit a sufficiently large non-linear reservoir to perform a basis expansion of the input into a high-dimensional latent space, enabling the downstream task to be more easily solved by the readout. This design circumvents the computational burden of traditional gradient descent optimization, and inherently avoids problems related to V/E gradients.

A fundamental baseline model in RC is the Leaky Echo State Network (LeakyESN), which exploits leaky-integrator neurons to tune the reservoir dynamics to the specific time scale of the input \cite{jaeger2007leakyesn}. The state transition function of a shallow LeakyESN is given by:
\begin{equation}
    \mathbf{h}(t) = (1 - \tau) \mathbf{h}(t-1) + \tau \phi \Big( \mathbf{W}_{h} \mathbf{h}(t-1) + \mathbf{W}_{x} \mathbf{x}(t) + \mathbf{b} \Big),
    \label{eq:leakyesn}
\end{equation}
where $\mathbf{h}(t) \in \mathbb{R}^{N_{h}}$ and $\mathbf{x}(t) \in \mathbb{R}^{N_{x}}$ are, respectively, the state and the external input at time step $t$. The recurrent weight matrix is denoted as $\mathbf{W}_{h} \in \mathbb{R}^{N_{h}\times N_{h}}$, the input weight matrix is denoted as $\mathbf{W}_{x} \in \mathbb{R}^{N_{h}\times N_{x}}$, $\mathbf{b} \in \mathbb{R}^{N_{h}}$ denotes the bias vector, $\phi$ denotes an element-wise applied non-linearity, and $\tau \in (0, 1]$ denotes the leaky rate hyperparameter. The entries of $\mathbf{W}_{x}$ are sampled randomly from a uniform distribution over $(-\omega_{x}, \omega_{x})$, where $\omega_{x}$ is the input scaling hyperparameter. Similarly, entries in $\mathbf{b}$ are sampled randomly from a uniform distribution over $(-\omega_{b}, \omega_{b})$, where $\omega_{b}$ represents the bias scaling hyperparameter. The entries in $\mathbf{W}_{h}$ are initially sampled from a uniform distribution over $(-1, 1)$ and subsequently rescaled to have a desired spectral radius $\rho$\footnote{The spectral radius of a matrix $\mathbf{A}$, sometimes denoted as $\rho(\mathbf{A})$, is defined as the largest among the lengths of its eigenvalues.}, a crucial hyperparameter governing the reservoir's dynamics and ESP. In practical applications, the spectral radius is generally constrained to be smaller than $1$.

The readout can be formulated as $\mathbf{y}(t) = \mathbf{W}_{o} \mathbf{h}(t)$, where $\mathbf{y}(t) \in \mathbb{R}^{N_{o}}$ denotes the network output at time step $t$ and $\mathbf{W}_{o} \in \mathbb{R}^{N_{o} \times N_{h}}$ denotes the readout weight matrix. The readout is typically trained using lightweight closed-form solutions, such as ridge regression or least squares methods.

Prior works have proposed more complex architectures to enhance reservoir expressivity. Deep Echo State Networks (DeepESNs) \cite{gallicchio2017deepesn} create a hierarchy of untrained reservoirs stacked on top of each other. DeepESNs provide an elementary modular organization of the dynamics of a reservoir system, and generalize the concept of shallow LeakyESNs towards deep architectural constructions. Edge of Stability Echo State Networks (ES$^{\text{2}}$N) \cite{ceni2024edge} introduced a class of ESNs in which the reservoir realizes a convex combination of a nonlinear transformation and a linear transformation modulated by a (random) orthogonal matrix. Residual Echo State Networks (ResESNs) \cite{ceni2024residual} extended ES$^{\text{2}}$N by introducing residual connections along the temporal dimension, modulated by various orthogonal transformations. In particular, ResESNs generalize the convex combination of traditional ESNs by relaxing it to a non-convex one, where the leaky rate $\tau$ and its complement $1 - \tau$ are replaced by independent scaling coefficients $\beta$ and $\alpha$, respectively. Euler State Networks (EuESN) \cite{gallicchio2024euler}, conceptually optimized for time series classification, leverage an Euler-like time step integration and a parametrization of the recurrent weight matrix based on an antisymmetric matrix.

\section{Deep Residual Echo State Networks}
\label{sec:deepresesn}
We introduce Deep Residual Echo State Networks (DeepResESNs), a novel class of deep untrained RNNs that merges the architectural principles of DeepESNs with the temporal residual connections of ResESNs. The proposed model consists of a hierarchy of untrained recurrent layers, where each layer benefits from a residual connection that propagates its previous state through a simple mapping, creating a shortcut along the temporal dimension. This design extends the advantages of residual connections to a deep, hierarchical context.

DeepResESN is characterized by a hierarchy of untrained recurrent layers based on temporal residual connections. Fig.~\ref{fig:deepresesn_architecture} graphically illustrates the proposed architecture. The first layer is driven by the external input, while subsequent layers are driven by the 
reservoir dynamics developed in the previous layer. 
To develop our mathematical description, let us assume a hierarchical construction of the reservoir with a number of $N_L$ layers, and let us use the superscript $(\cdot)^{(l)}$ to generally refer to the network's weights and hyperparameters at layer $l$, for $l = 1, \ldots, N_{L}$. The state transition function computed by the $l$-th layer of a DeepResESN is given by:
\begin{equation}
\mathbf{h}^{(l)}(t) = \alpha^{(l)} \mathbf{O} \mathbf{h}^{(l)}(t - 1) +\beta^{(l)} \phi ( \mathbf{W}_h^{(l)} \mathbf{h}^{(l)}(t - 1) + \mathbf{W}_x^{(l)} \mathbf{x}^{(l)}(t) + \mathbf{b}^{(l)} ),
\label{eq:deepresesn}
\end{equation}
where the input is given by the external signal for $l= 1$, i.e., $\mathbf{x}^{(1)}(t) = \mathbf{x}(t) $, and by the reservoir state of the previous layer for $l>1$, i.e.,  $\mathbf{x}^{(l)}(t) = \mathbf{h}^{(l-1)}(t)$. Assuming, for simplicity, that all reservoir layers have the same hidden size $N_{h}$, we have $\mathbf{W}_x^{(1)} \in \mathbb{R}^{N_{h}\times N_{x}}$ for $l=1$, and $\mathbf{W}_x^{(l)} \in \mathbb{R}^{N_{h}\times N_{h}}$ for $l>1$. Here, $\mathbf{O} \in \mathbb{R}^{N_{h}\times N_{h}}$ is an orthogonal matrix, and $\alpha^{(l)} \in [0, 1]$ and $\beta^{(l)} \in (0, 1]$ are layer-specific scaling coefficients that generalize the leaky rate mechanism from \eqref{eq:leakyesn}. In addition, each layer $l$ may have its own set of hyperparameters $\rho^{(l)}$, $\omega_x^{(l)}$, and $\omega_b^{(l)}$, all employed in the same way as previously described for LeakyESN. In this work we consider the hyperbolic tangent as the element-wise applied non-linearity (i.e., $\phi = \tanh$). The readout may be fed either the states from the last reservoir layer $\mathbf{h}^{(N_{L})}$ or the concatenation of states from all reservoir layers $[\mathbf{h}^{(1)}, \mathbf{h}^{(2)}, \ldots, \mathbf{h}^{(N_{L})}]$, as shown in Fig.~\ref{fig:deepresesn_architecture}.

\begin{figure}[!t]
    \includegraphics[scale=1.0, center]{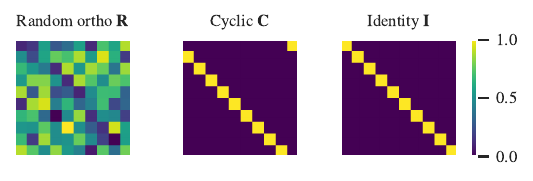}
    \caption{Structure of the three orthogonal matrices ($10 \times 10$) used in the temporal residual connections.}
    \label{fig:orthogonal_matrices}
\end{figure}

We consider three types of orthogonal matrices for the temporal residual connections in DeepResESN, one randomly generated and two with a fixed structure (see Fig.~\ref{fig:orthogonal_matrices} for a graphical representation). The first, denoted by $\mathbf{R}$, is obtained via QR decomposition of a random $N_{h} \times N_{h}$ matrix with real-valued i.i.d entries in $[-1, 1)$. The second one, denoted by $\mathbf{C}$, is the cyclic orthogonal matrix \cite{rodan2010minimum, tino2020dynamical}, a zero matrix with ones on the sub-diagonal and in the top-right corner. The third one is the identity matrix, denoted by $\mathbf{I}$. The choice of these three cases is motivated by their distinct impact on short-term memory capacity, covering a broad spectrum of dynamical behaviors \cite{jaeger2002memory}. In particular, the identity matrix provides relatively low memory capacity while the two other orthogonal matrices maximize it \cite{ceni2024edge}. Additionally, the random orthogonal matrix offers a stochastic scenario due to its random generation, and the cyclic orthogonal matrix provides a deterministic alternative due to its fixed structure. These configurations thus provide a representative set of dynamical regimes for analyzing the impact of different residual connectivity structures in DeepResESN.

The proposed DeepResESN generalizes several existing ESN architectures. DeepESN, whose per-layer dynamics are governed by a leaky reservoir as in \eqref{eq:leakyesn}, corresponds to a DeepResESN where the temporal residual connections employ the identity matrix $\mathbf{I}$ and $\beta = 1 - \alpha$. ResESN corresponds to a DeepResESN with a single reservoir layer, $N_{L} = 1$. ES$^{\text{2}}$N and EuESN dynamics can also be derived from the per-layer dynamics of the DeepResESN framework in \eqref{eq:deepresesn}. Specifically, assuming $N_{L} = 1$, ES$^{\text{2}}$N corresponds to a DeepResESN employing a random orthogonal matrix $\mathbf{R}$ with $\beta = 1 - \alpha$. EuESN is obtained by setting $\alpha = 1$ and $\beta = \epsilon$, where $\epsilon > 0$ is the Euler integration step hyperparameter, and by parameterizing $\mathbf{W}_h = \mathbf{A} - \gamma \mathbf{I}$, where $\gamma \in [0, 1)$ is the diffusion hyperparameter and $\mathbf{A}$ is an antisymmetric matrix.

\begin{figure}[!b]
    \includegraphics[scale=1.0, center]{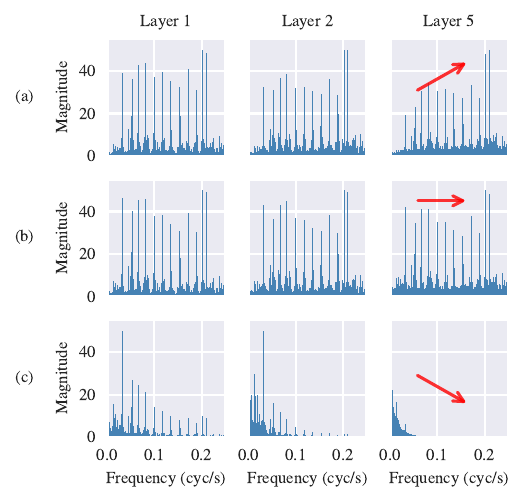}
    \caption{Spectral frequencies of (a) DeepResESN$_\mathrm{R}$, (b) DeepResESN$_\mathrm{C}$, and (c) DeepResESN$_\mathrm{I}$, in progressively deeper layers (columns). In each layer, and for all configurations, we consider $N_{h} = 100$ recurrent neurons, $\rho = 1$, $\alpha = 0.9$, $\beta = 0.1$, $\omega_x = 1$ and $\omega_b = 0$. Results are averaged over $10$ trials. Magnitudes have been normalized to ease visualization. Red arrows highlight the trend in spectral magnitudes.}
    \label{fig:spectral}
\end{figure}

\subsection{Spectral Frequency Analysis}
\label{subsec:spectral}
Here, leveraging spectral frequency analysis tools, we investigate how the temporal representation of the input signal is encoded in progressively deeper layers of DeepResESN. Consider the time series $\mathbf{s}(t) = \sum_{i=1}^{12} \sin(\varphi_i  t)$, where $\varphi_i$ is the $i$-th frequency in $\varphi = [0.2, 0.331, 0.42, 0.51, 0.63, 0.74, 0.85, 0.97, 1.08, 1.19, 1.27, 1.32]$. Frequency values are the same as in \cite{koryakin2012balanced, otte2016optimizing}. The experiment can be summarized in three steps: (i) generate a time series $\mathbf{s}(t)$ of length $T = 1000$; (ii) feed it to a DeepResESN and collect the output hidden states at each layer; and (iii) apply a Fast Fourier Transform (FFT)\cite{frigo98fft} to extract the frequency components over time. Fig.~\ref{fig:spectral} presents the frequency components at progressively deeper layers for each DeepResESN configuration. 

In Fig.~\ref{fig:spectral}~(c), we observe that the identity configuration tends to filter out higher frequencies, with the filtering effect becoming stronger in deeper layers. In contrast, the random orthogonal configuration (Fig.~\ref{fig:spectral}~(a)) tends to filter out lower frequencies, while the cyclic orthogonal configuration (Fig.~\ref{fig:spectral}~(b)) appears to maintain frequencies relatively unchanged regardless of layer depth. These distinct behaviors highlight the importance of exploring different configurations for the temporal residual connections, as the temporal representation of the input can change depending on the specific configuration, particularly in deeper layers.

\section{Stability Analysis}
\label{sec:mathematics}
The dynamics of randomized, non-linear dynamical systems are crucial for assessing their effectiveness in learning and capturing relevant information. Thus, the study of such dynamics through appropriate mathematical tools may provide valuable insights into the system's behavior and operating regimes. In this section, we extend the ESP to DeepResESN and study its dynamics through the lens of linear stability analysis.

Due to its hierarchical structure, the dynamics of layer $l$ in a DeepResESN depend on the transformations of all previous layers and can therefore be expressed in terms of their collective states. Let us denote with 
$ \{\mathbf{h}^{(k)}(t-1)\}_{k=1}^{l} = \{ \mathbf{h}^{(1)}(t-1), \ldots, \mathbf{h}^{(l)}(t-1) \}  $
the collection of the states of all layers up to layer $l$ (inclusive) at time step $t - 1$. For the first layer $(l = 1)$, state $\mathbf{h}^{(1)}(t)$ depends on the external input $\mathbf{x}(t)$ and on $\mathbf{h}^{(1)}(t -1)$. For layers after the first ($l > 1$), state $\mathbf{h}^{(l)}(t)$ depends on $\mathbf{h}^{(l - 1)}(t)$, which can be expressed as a function of $\mathbf{x}(t)$ and $\{\mathbf{h}^{(k)}(t-1)\}_{k=1}^{l-1}$, and on $\mathbf{h}^{(l)}(t -1)$. Therefore, in general, state $\mathbf{h}^{(l)}(t)$ depends on the states of all layers up to $l$ (inclusive) at time step $t - 1$.  We can capture these dependencies by defining a function $F^{(l)}$, the state transition function computed by the $l$-th layer, that encapsulates the network's dynamics up to layer $l$:
\begin{align*}
F^{(l)} : \mathbb{R}^{N_x} \times \underbrace{\mathbb{R}^{N_h} \times \ldots \times \mathbb{R}^{N_h}}_\text{$l$} \rightarrow \mathbb{R}^{N_h}
\end{align*}
\begin{align}
&\mathbf{h}^{(l)}(t) = F^{(l)} ( \mathbf{x}(t), \{\mathbf{h}^{(k)}(t-1)\}_{k=1}^{l} ) 
\label{eq:deepresesnFi}
\end{align}

Then, denoting with $F = (F^{(1)},  \ldots,  F^{(N_L)})$ the function that expresses the global dynamics of a DeepResESN, the evolution of the system reads:
\begin{equation*}
F : \mathbb{R}^{N_x} \times \underbrace{\mathbb{R}^{N_h} \times \ldots \times \mathbb{R}^{N_h}}_\text{$N_L$} \rightarrow \underbrace{\mathbb{R}^{N_h} \times \ldots \times \mathbb{R}^{N_h}}_\text{$N_L$}
\end{equation*}
\begin{equation}
\mathbf{h}(t) = F \big( \mathbf{x}(t),  \mathbf{h}(t-1) \big).
\label{eq:deepresesnF} 
\end{equation}

Before delving deeper, let us define function $\hat{F}$ to denote the iterated version of the global state transition function $F$ defined in \eqref{eq:deepresesnF}. Assume an input sequence of length $T$, denoted as $\mathbf{s}_{T} \in (\mathbb{R}^{N_x})^{*}$ and global state $\mathbf{h} \in \mathbb{R}^{N_L N_h}$. Then, $\hat{F}(\mathbf{s}_{T}, \mathbf{h})$ is the global state of a DeepResESN driven by external input $\mathbf{s}_{T}$ with initial state $\mathbf{h}$. More formally:
\begin{equation*}
\hat{F}  :  (\mathbb{R}^{N_x})^{*} \times \underbrace{\mathbb{R}^{N_h} \times \ldots \times \mathbb{R}^{N_h}}_\text{$N_L$} \rightarrow \underbrace{\mathbb{R}^{N_h} \times \ldots \times \mathbb{R}^{N_h}}_\text{$N_L$}
\end{equation*}
\begin{equation}
{\hat{F}(\mathbf{s}_{T}, \mathbf{h})} = 
\begin{cases}
\mathbf{h}, &{\mathrm{if}}\ \mathbf{s}_{T} = [ \; ]\ \mathrm{(null\ seq.)}, \\
F \big( \mathbf{x}_{T}, \hat{F}(\{\mathbf{x}(t)\}_{t=1}^{T-1}, \mathbf{h}) \big), &{\mathrm{if}}\ \mathbf{s}_{T} = (\mathbf{x}(1), \ldots, \mathbf{x}(T)).
\end{cases}
\label{eq:deepresesnF_iterated}
\end{equation}

For the ESP of DeepResESN we use the definition introduced in \cite{gallicchio2017deepesn} for hierarchical ESNs. We require that, for any input sequence, the effect of initial conditions vanishes asymptotically. Assume a DeepResESN whose global dynamics are governed by function $F$ defined in \eqref{eq:deepresesnF}. Then, the ESP holds if for any input sequence of length $T$, $\mathbf{s}_{T} = [\mathbf{x}(1),  \ldots , \mathbf{x}(T)]$, and for all pairs of initial states $\mathbf{h}, \mathbf{h}' \in \mathbb{R}^{N_L N_h}$ it holds that:
\begin{equation}
    \lim_{T \rightarrow \infty}\lVert \hat{F}(\mathbf{s}_{T}, \mathbf{h}) - \hat{F}(\mathbf{s}_{T}, \mathbf{h}') \rVert = 0.
    \label{eq:deepresesnESP}
\end{equation}

\subsection{Stability of DeepResESN dynamics}
Here, we provide a necessary condition for the ESP of DeepResESN. Following standard practice in the RC literature, we consider a linearized version of the system expressed in \eqref{eq:deepresesnF}. We obtain this linearization via a first-order Taylor approximation evaluated around a point $\mathbf{h}_{0} \in \mathbb{R}^{N_L N_h}$. The linearized system reads:
\begin{equation}
    \begin{aligned}
    \mathbf{h}(t) 
    & \: = \: \mathbf{J}_{F, \mathbf{h}}(\mathbf{x}(t), \mathbf{h}_{0})  (\mathbf{h}(t-1) - \mathbf{h}_{0}) + F(\mathbf{x}(t), \mathbf{h}_{0}),
    \end{aligned}
    \label{eq:deepresesnF_linearized}
\end{equation}
where $\mathbf{J}_{F, \mathbf{h}}(\mathbf{x}(t), \mathbf{h}_{0})$ is the Jacobian of DeepResESN, evaluated at $\mathbf{h}_{0}$ and with external input $\mathbf{x}(t)$. Given the hierarchical structure of the model, $\mathbf{J}_{F, \mathbf{h}}(\mathbf{x}(t), \mathbf{h}_{0})$ can be expressed in terms of the Jacobians of each layer, which we refer to as inter-layer Jacobians. This results in the Jacobian being a block matrix. For every $i, j = 1, \ldots, N_L$, the inter-layer Jacobian of $F^{(i)}$, with respect to state $\mathbf{h}^{(j)}(t-1)$ and evaluated at $\mathbf{h}_{0}$, is given by:
\begin{equation}
    \mathbf{J}_{F^{(i)}, \mathbf{h}^{(j)}}(\mathbf{x}(t), \mathbf{h}_{0}) = 
    \begin{cases}
        \frac{\partial F^{(i)} ( \mathbf{x}(t),  \{\mathbf{h}_{0}^{(k)}(t-1)\}_{k=1}^{i} ) }{\partial \mathbf{h}_{0}^{(j)}(t-1)}, & \text{if}\ i \geq j, \\
        \mathbf{0}, & \text{otherwise}
    \end{cases}
    \label{eq:deepresesnFi_jacobian}
\end{equation}
Note that the partial derivative in \eqref{eq:deepresesnFi_jacobian} is a zero matrix whenever $i < j$, since the state transition function $F^{(i)}$ at layer $i$ depends on the states of layers up to and including $i$ and is independent of the states in subsequent ones. For details on its structure, see Appendix~\ref{subsec:proof_necessary_esp} and \eqref{eq:diagonal_block_matrices} in particular. The Jacobian of a DeepResESN can now be expressed as follows:
\begin{align}
& \mathbf{J}_{F, \mathbf{h}}(\mathbf{x}(t), \mathbf{h}_{0}) \notag \\[0.2cm]
& = 
\left( \begin{array}{lll}
    \mathbf{J}_{F^{(1)}, \mathbf{h}^{(1)}}(\mathbf{x}(t), \mathbf{h}_{0})   & \cdots & \mathbf{J}_{F^{(1)}, \mathbf{h}^{(N_L)}}(\mathbf{x}(t), \mathbf{h}_{0}) \\
    \quad\quad\vdots & \ddots & \quad\quad\vdots\\
    \mathbf{J}_{F^{(N_L)}, \mathbf{h}^{(1)}}(\mathbf{x}(t), \mathbf{h}_{0}) & \cdots & \mathbf{J}_{F^{(N_L)}, \mathbf{h}^{(N_L)}}(\mathbf{x}(t), \mathbf{h}_{0})
\end{array} \right) \notag \\[0.2cm]
& \overset{\text{$\ast$}}{=} 
\left( \begin{array}{llc}
    \mathbf{J}_{F^{(1)}, \mathbf{h}^{(1)}}(\mathbf{x}(t), \mathbf{h}_{0}) & & \multirow{2}{*}{\text{\huge 0}} \\
    \quad\quad\vdots & \ddots & \\
    \mathbf{J}_{F^{(N_L)}, \mathbf{h}^{(1)}}(\mathbf{x}(t), \mathbf{h}_{0}) & \cdots & \mathbf{J}_{F^{(N_L)}, \mathbf{h}^{(N_L)}}(\mathbf{x}(t), \mathbf{h}_{0})
\end{array} \right),
\label{eq:deepresesnF_jacobian}
\end{align}
where the step marked by $\ast$ follows from \eqref{eq:deepresesnFi_jacobian} and the fact that state transition function $F^{(i)}$ at layer $i$ depends on the states of layers up to and including $i$. The linearized system in \eqref{eq:deepresesnF_linearized} is (asymptotically) stable if and only if, in the case of zero input, the spectral radius of the Jacobian, denoted by $\rho(\mathbf{J})$, is strictly less than one. The spectral radius of the Jacobian plays a crucial role in determining the regime under which a linear dynamical system operates. The system (without input) is stable when the magnitudes of all the eigenvalues of $\mathbf{J}$ are smaller than $1$, as the system will asymptotically approach an equilibrium point. Conversely, if any eigenvalue has a magnitude greater than $1$, the system may exhibit instability. 

\begin{theorem}[Necessary condition for the ESP of a DeepResESN]
\label{th:necessary_esp}
Assume a DeepResESN whose inter-layer and global dynamics are defined in \eqref{eq:deepresesnFi} and \eqref{eq:deepresesnF}, respectively. Furthermore, assume zero input and zero initial state. The global spectral radius of the system is expressed as:
\begin{equation}
    \rho(\mathbf{J}_{F, \mathbf{h}} (\mathbf{0}_x, \mathbf{0})) = \max_{l = 1, \ldots, N_L} \rho ( \alpha^{(l)}\mathbf{O} + \beta^{(l)}\mathbf{W}_{h}^{(l)} ).
    \label{eq:deepresesn_global_rho}
\end{equation}
where $\mathbf{0}_x \in \mathbb{R}^{N_x}$ and $\mathbf{0} \in \mathbb{R}^{N_L N_h}$ are the zero input and the zero state vectors, respectively. Then, a necessary condition for the ESP of a DeepResESN is given by:
\begin{equation}
    \rho(\mathbf{J}_{F, \mathbf{h}} (\mathbf{0}_x, \mathbf{0})) < 1.
    \label{eq:deeresesn_necessaryESP}
\end{equation}
\end{theorem}
\noindent The proof is given in Appendix~\ref{subsec:proof_necessary_esp}.

\subsection{Contractivity of  dynamics}
We provide a sufficient condition for the ESP of DeepResESN. In RC systems, this is generally established by studying the conditions under which reservoir dynamics act as a contraction mapping. This implies that, for any input sequence, the distance between two hidden states decreases at each time step by a factor of $0<C<1$, where $C$ is the coefficient of contraction at each time step\footnote{For a generic state transition function $F$, any input $\mathbf{x}$, and states $\mathbf{h}$, $\mathbf{h}'$: $\|F(\mathbf{x}, \mathbf{h}) - F(\mathbf{x}, \mathbf{h}')\|\leq C \|\mathbf{h}-\mathbf{h}'\|$, with $C\in (0,1)$.}. We adopt the contractivity framework of \cite{gallicchio2017deepesnesp}, including the Euclidean distance (or L2 norm) to express the distance between individual layers' states, the maximum product metric for the distance between global states, and layer-wise and global contractivity definitions.

\begin{lemma}[Sufficient condition for contractivity of layer's dynamics]
\label{lemma:layer_contractivity}
Consider a DeepResESN whose dynamics at layer $l$ are expressed by the state transition function $F^{(l)}$, defined in \eqref{eq:deepresesnFi}. Then, a sufficient condition for the contractivity of the dynamics of the reservoir at layer $l$ is given by:
\begin{equation}
    C^{(l)} = \alpha^{(l)} + \beta^{(l)} ( \lVert \mathbf{W}^{(l)}_h \rVert + C^{(l-1)} \lVert \mathbf{W}^{(l)}_x \rVert ) < 1,
    \label{eq:contractivity_layer}
\end{equation}
assuming $F^{(l-1)}$ to be contractive with coefficient $C^{(l-1)} < 1$. Furthermore, assuming $C^{(0)} = 0$, for the first layer ($l = 1$), the input weight matrix contribution cancels out and \eqref{eq:contractivity_layer} simplifies to $C^{(1)} = \alpha^{(1)} + \beta^{(1)} \lVert \mathbf{W}^{(1)}_h \rVert < 1$.
\end{lemma}
\noindent The proof is given in Appendix~\ref{subsec:proof_layer_contractivity}.

\begin{theorem}[Sufficient condition for the ESP of a DeepResESN]
\label{th:sufficient_esp}
Consider a DeepResESN whose layer dynamics and global dynamics are expressed by state transition functions $F^{(i)}$ in \eqref{eq:deepresesnFi} and $F$ in \eqref{eq:deepresesnF}, respectively. 
Assume that for every $l = 1, \ldots, N_L$ the dynamics at layer $l$ are contractive with a coefficient of contraction $C^{(l)}$. Furthermore, assume the DeepResESN to have global dynamics that implement a contraction with coefficient $C$, where:
\begin{equation}
    C = \max_{l = 1, \ldots, N_L} C^{(l)} < 1.
    \label{eq:contractivity_global}
\end{equation}

Then,  assuming the reservoir global state space is bounded, the DeepResESN satisfies the ESP for all inputs.
\end{theorem}
\noindent The proof is given in Appendix~\ref{subsec:proof_sufficient_esp}.

\begin{figure}[!b]
    \includegraphics[scale=1.0, center]{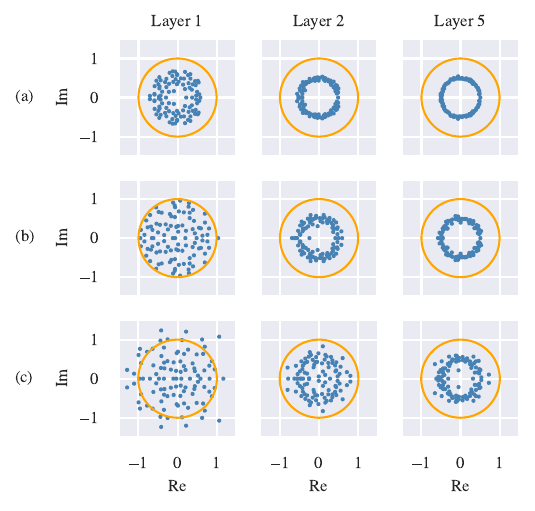}
    \caption{Eigenvalues of the Jacobian of a DeepResESN$_{\mathrm{R}}$ for spectral radii (a) $\rho = 0.5$, (b) $\rho = 1$, and (c) $\rho = 2$, for progressively deeper layers (columns). In each layer, we consider $N_{h} = 100$ recurrent neurons, $\rho$ as specified in each subplot, $\alpha = 0.5$, $\beta = 1$, $\omega_x = 1$, and $\omega_b = 0$. Model dynamics are driven by a random input vector and a random hidden state, both uniformly distributed in $(-1, 1)$. In orange the unitary circle.}
    \label{fig:eigenvalues}
\end{figure}

\subsection{Eigenspectrum Analysis}
\label{eigenspectrum}
The eigenspectrum\footnote{The eigenspectrum of a matrix refers to the set of all its eigenvalues.} of the Jacobian can reveal fundamental properties about the network's underlying behavior and dynamics. In particular, eigenvalues with magnitudes greater than one indicate a potentially chaotic regime, while eigenvalues with magnitudes smaller than one suggest asymptotic stability. In this section, we analyze the eigenspectrum of a DeepResESN$_{\mathrm{R}}$ for progressively higher values of the spectral radius $\rho$ (rows) and across progressively deeper layers (columns). This analysis is illustrated in Fig.~\ref{fig:eigenvalues}. We observe diverse eigenspectra across different depths (Figures~\ref{fig:eigenvalues}(a) and (b)), suggesting that deeper networks exhibit richer dynamics due to different layers being characterized by distinct eigenspectra. Notably, the eigenvalues tend to concentrate more around the origin as the network deepens. Additionally, Fig.~\ref{fig:eigenvalues}(c) reveals a stabilization effect in deeper layers: while eigenvalues fall outside the unit circle in the first layer, they are fully contained within the unit circle in deeper ones.

\section{Experiments}
\label{sec:experiments}
We validate the proposed approach across memory-based, forecasting, and classification tasks for time series.

\begin{table}[!b]
\caption{Model selection hyperparameters.}
\begin{center}
\begin{threeparttable}
\begin{tabular}{ll}
\toprule
Hyperparameters & Values \\
\midrule
concat & [False, True] \\
$N_L$ & [2, 3, 4, 5] \\
$\omega_x$ & [0.01, 0.1, 1, 10] \\
$\omega_b$ & [0, 0.01, 0.1, 1, 10] \\ 
\midrule
\textbf{(Leaky variants)} & \\
$\rho$ & [0.9, 1, 1.1] \\
$\tau$ & [0.0001, 0.1, 0.5, 0.9, 0.99, 1]\tnote{\textcolor{cyan}{*}} \\
\midrule
\textbf{(Residual variants)} & \\
$\rho$ & [0.9, 1, 1.1] \\
$\alpha$ & [0, 0.0001, 0.1, 0.5, 0.9, 0.99, 1]\tnote{\textcolor{cyan}{*}} \\
$\beta$ & [0.0001, 0.1, 0.5, 0.9, 0.99, 1]\tnote{\textcolor{cyan}{*}} \\
\midrule
\textbf{(Euler variants)} & \\
$\omega_{r}$ & [0.01, 0.1, 1, 10] \\
$\epsilon$ & [0.0001, 0.1, 0.5, 0.9, 0.99, 1]\tnote{\textcolor{cyan}{*}} \\
$\gamma$ & [0, 0.1, 0.5, 0.9, 1] \\
\bottomrule
\end{tabular}
\begin{tablenotes}
\item[\textcolor{cyan}{*}] Values $0.0001$ and $0.99$ have been explored only for memory-based tasks.
\end{tablenotes}
\label{tab:model_selection}
\end{threeparttable}
\end{center}
\end{table}

\begin{figure*}[t]
    \centering
    \begin{minipage}[t]{\textwidth}    
        \begin{minipage}[b]{0.5\textwidth}
            \centering
            \captionsetup[subfloat]{width=\linewidth}
            \subfloat{\includegraphics[scale=1.0, left]{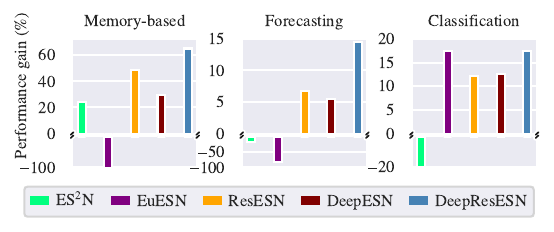}
            \label{fig:performance_change}}
        \end{minipage}\hfill
        \begin{minipage}[b]{0.5\textwidth}
            \centering
            \captionsetup[subfloat]{width=\linewidth}
            \subfloat{\includegraphics[scale=1.0, right]{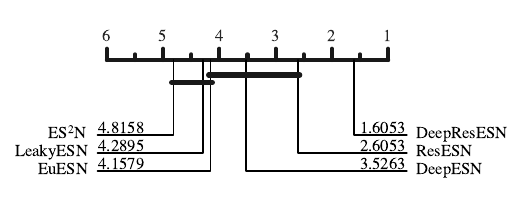}
            \label{fig:critical_difference}}
        \end{minipage}\hfill
    \end{minipage}
    \caption{(\emph{left}) Performance gain of each model class relative to LeakyESN, broken down by task and averaged across all task-specific datasets. For ResESNs and DeepResESNs, we consider the best-performing configuration for each dataset. (\emph{right}) Critical difference plot computed via a Wilcoxon test \cite{demvsar2006statistical}, summarizing the average rank (lower is better) of each model class across all tasks and datasets. Cliques, represented as solid lines, connect models with no statistically significant difference in performance.}
    \label{fig:performance_summary}
\end{figure*}

\subsubsection*{Experimental setting}
Model selection is carried out via Bayesian search. For each model, we explore up to $1000$ configurations with a maximum runtime of $24$h. Explored hyperparameters are highlighted in Table~\ref{tab:model_selection}. Results are averaged across $10$ random initializations and the best configuration is chosen based on the performance achieved on the validation set. Hyperparameters common to all models include the input scaling $\omega_{x}$ and the bias scaling $\omega_{b}$. For leaky variants, LeakyESN, ES$^{\text{2}}$N, and DeepESN, we explore the leaky rate $\tau$. For residual variants, ResESN and DeepResESN, we explore the coefficients $\alpha$ and $\beta$. For both leaky and residual variants, we also explore the spectral radius $\rho$ used to rescale the recurrent weight matrix $\mathbf{W}_{h}$. For Euler variants, we explore the recurrent scaling $\omega_{r}$ used to rescale the antisymmetric matrix, the Euler integration step $\epsilon$, and the diffusion term $\gamma$. For deep models, DeepESN and DeepResESN, we also explore the number of layers $N_{L}$ and an additional set of hyperparameters for layers beyond the first one. We also consider hyperparameter $\texttt{concat} \in [\text{False}, \text{True}]$, which determines whether the readout is fed the states of the last layer or the concatenation of the states across all layers, respectively (see Fig.~\ref{fig:deepresesn_architecture}(b)). To ensure a consistent number of trainable parameters when states are concatenated, the hidden size (i.e., the total number of reservoir units) is evenly split across layers. If an even split is not possible, the remaining neurons are allocated to the first layer. All models employ fully-connected reservoirs consisting of $N_{h} = 100$ recurrent neurons. The readout is trained via ridge regression with a Singular Value Decomposition (SVD) solver. The regularization coefficient $\lambda$ of the readout is fixed at $0$ for memory-based and forecasting tasks, while for classification tasks, we explore values in $[0, 0.01, 0.1, 1, 10, 100]$.

\subsubsection*{Overall performance analysis}
Fig.~\ref{fig:performance_summary} provides an overview of DeepResESN's competitiveness relative to traditional shallow and deep RC, presenting the average performance change relative to a LeakyESN baseline (left) and a critical difference plot ranking models based on their overall performance (right). In the left panel, we observe that DeepResESN delivers substantial improvements over a traditional LeakyESN, achieving performance gains of approximately $+65.1\%$, $+14.4\%$, and $+17.5\%$ for memory-based, forecasting, and classification tasks, respectively. These improvements consistently exceed those provided by ES$^{\text{2}}$N, ResESNs, and DeepESNs. While EuESN performs slightly better than DeepResESN in classification, the trade-off is substantially degraded performance on memory-based and forecasting tasks, where EuESN is outperformed by LeakyESN by roughly $100\%$ and $80\%$, respectively. The right panel further highlights DeepResESN's advantages. Our model ranks first in terms of overall performance. Notably, this advantage is statistically significant, as evidenced by the absence of any clique connecting to DeepResESN. See Sections~\ref{sec:memory_experiments}, \ref{sec:forecasting_experiments}, and \ref{sec:classification_experiments} for details on memory-based, forecasting and classification results, respectively.

\subsection{Memory-based}
\label{sec:memory_experiments}
The memory-based tasks considered are ctXOR, a variant of the XOR task introduced in \cite{verstraeten2010ctxor}, and SinMem, inspired by the function approximation task from \cite{inubushi2017sinmem}.

\subsubsection*{Methodology} We use the Normalized Root Mean Squared Error (NRMSE) as the target metric. We generate a time series of length $T = 6000$, with the first $4000$ time steps used for training, the next $1000$ for validation, and the final $1000$ for testing. Training and inference phases employ a $200$ time step washout to warm up the reservoir. The readout is fed the states at each time step.

\subsubsection*{ctXOR} The ctXOR task evaluates the model's trade-off between memorization and non-linear processing capabilities. Consider a one-dimensional input time series $\mathbf{x}(t)$ uniformly distributed in $(-0.8, 0.8)$ and assume $\mathbf{r}(t - d) = \mathbf{x}(t - d - 1)  \mathbf{x}(t - d)$. The task is to output the time series $\mathbf{y}(t) = \mathbf{r}(t - d)^{p}  \sign(\mathbf{r}(t - d))$, where $d$ is the delay and $p$ determines the strength of the non-linearity. We consider delay $d = 5$ (ctXOR5) and delay $d = 10$ (ctXOR10). The non-linearity strength is set to $p = 2$.

\subsubsection*{SinMem} The SinMem task tests the model's ability to reconstruct a non-linear transformation of a past version of the input. Given a one-dimensional input time series $\mathbf{x}(t)$ uniformly distributed in $(-0.8, 0.8)$, the task is to output the time series $\mathbf{y}(t) = \sin ( \pi  \mathbf{x}(t - d) )$. For SinMem, we consider delays $d = 10$ (SinMem10) and $d = 20$ (SinMem20).

\subsubsection*{Discussion} Table~\ref{tab:memory_results} presents the results for memory-based tasks. DeepResESN consistently outperforms shallow and deep RC on all tasks. In particular, the performance gap increases when greater long-term memorization capabilities are required due to bigger delays. Notably, EuESN lacks strong memorization capabilities, resulting in errors up to nearly three orders of magnitude higher than those of DeepResESN on the SinMem tasks. Finally, performance is heavily influenced by the specific configuration employed in the temporal residual connections. Those employing either a random orthogonal or a cyclic orthogonal matrix considerably outperform those employing the identity matrix, leading to a lower error by approximately one order of magnitude in SinMem10 and SinMem20 tasks.

\begin{table*}[t]
\caption{Test set results on memory-based tasks, averaged over $10$ different random initializations. The \textbf{best result} is in bold.}
\label{tab:memory_results}
\centering
\resizebox{\textwidth}{!}{
\footnotesize
\begin{NiceTabular}{@{}l | c c c c c c c | c c c@{}}
\toprule
$\downarrow$ Memory-based & \rowcolor{baseline} LeakyESN & ES$^{\text{2}}$N & EuESN & ResESN$_{\mathrm{R}}$ & ResESN$_{\mathrm{C}}$ & ResESN$_{\mathrm{I}}$ & DeepESN & \rowcolor{our} DeepResESN$_{\mathrm{R}}$ & DeepResESN$_{\mathrm{C}}$ & DeepResESN$_{\mathrm{I}}$ \\
\midrule
\textsc{ctXOR5} $(\cdot 10^{-1})$       & \rowcolor{baseline} $3.6_{\pm 0.1}$ & $3.6_{\pm 0.1}$ & $10.0_{\pm 0.1}$ & $3.6_{\pm 0.1}$ & $3.6_{\pm 0.1}$ & $3.6_{\pm 0.1}$ & $3.7_{\pm 0.1}$ & \rowcolor{our} $\mathbf{3.2_{\pm 0.2}}$ & $3.6_{\pm 0.0}$ & $3.6_{\pm 0.0}$ \\

\textsc{ctXOR10} $(\cdot 10^{-1})$      & \rowcolor{baseline} $8.7_{\pm 0.3}$ & $9.1_{\pm 0.5}$ & $10.1_{\pm 0.1}$ & $7.3_{\pm 0.9}$ & $6.6_{\pm 1.0}$ & $8.5_{\pm 0.4}$ & $6.5_{\pm 1.0}$ & \rowcolor{our} $\mathbf{4.1_{\pm 0.2}}$ & $4.8_{\pm 0.5}$ & $6.2_{\pm 0.7}$ \\

\textsc{SinMem10} $(\cdot 10^{-2})$     & \rowcolor{baseline} $35.9_{\pm 0.2}$ & $15.4_{\pm 0.4}$ & $72.0_{\pm 0.6}$ & $0.2_{\pm 0.0}$ & $0.2_{\pm 0.0}$ & $36.3_{\pm 0.2}$ & $14.8_{\pm 1.7}$ & \rowcolor{our} $0.5_{\pm 0.1}$ & $\mathbf{0.1_{\pm 0.0}}$ & $0.8_{\pm 0.1}$ \\

\textsc{SinMem20} $(\cdot 10^{-2})$     & \rowcolor{baseline} $37.6_{\pm 0.1}$ & $21.4_{\pm 1.2}$ & $80.8_{\pm 0.2}$ & $12.7_{\pm 1.4}$ & $11.3_{\pm 1.3}$ & $37.6_{\pm 0.2}$ & $24.0_{\pm 3.8}$ & \rowcolor{our} $1.9_{\pm 0.5}$ & $\mathbf{1.2_{\pm 1.8}}$ & $24.7_{\pm 2.5}$ \\
\bottomrule
\end{NiceTabular}
}
\end{table*}

\begin{table*}[t]
\caption{Test set results on forecasting tasks, averaged over $10$ different random initializations. The \textbf{best result} is in bold.}
\label{tab:forecasting_results}
\centering
\resizebox{\textwidth}{!}{
\footnotesize
\begin{NiceTabular}{@{}l | c c c c c c c | c c c@{}}
\toprule
$\downarrow$ Forecasting & \rowcolor{baseline} LeakyESN & ES$^{\text{2}}$N & EuESN & ResESN$_{\mathrm{R}}$ & ResESN$_{\mathrm{C}}$ & ResESN$_{\mathrm{I}}$ & DeepESN & \rowcolor{our} DeepResESN$_{\mathrm{R}}$ & DeepResESN$_{\mathrm{C}}$ & DeepResESN$_{\mathrm{I}}$ \\
\midrule
\textsc{Lz25} $(\cdot  10^{-2})$    & \rowcolor{baseline} $13.9_{\pm 0.4}$ & $12.7_{\pm 0.3}$ & $18.7_{\pm 0.2}$ & $12.2_{\pm 0.5}$ & $\mathbf{11.6_{\pm 0.4}}$ & $12.7_{\pm 0.2}$ & $14.7_{\pm 0.6}$ & \rowcolor{our} $12.2_{\pm 0.2}$ & $12.2_{\pm 0.5}$ & $12.6_{\pm 0.4}$ \\

\textsc{Lz50} $(\cdot  10^{-2})$    & \rowcolor{baseline} $33.7_{\pm 0.6}$ & $34.0_{\pm 0.4}$ & $41.4_{\pm 0.4}$ & $33.9_{\pm 0.6}$ & $34.4_{\pm 0.5}$ & $33.0_{\pm 0.3}$ & $34.6_{\pm 1.2}$ & \rowcolor{our} $32.2_{\pm 0.5}$ & $\mathbf{32.0_{\pm 0.6}}$ & $33.5_{\pm 0.6}$ \\

\textsc{MG} $(\cdot  10^{-4})$      & \rowcolor{baseline} $3.3_{\pm 0.3}$ & $3.3_{\pm 0.2}$ & $15.1_{\pm 0.8}$ & $3.1_{\pm 0.3}$ & $3.2_{\pm 0.3}$ & $3.2_{\pm 0.3}$ & $3.1_{\pm 0.4}$ & \rowcolor{our} $\mathbf{3.0_{\pm 0.2}}$ & $3.2_{\pm 0.3}$ & $3.2_{\pm 0.2}$ \\

\textsc{MG84} $(\cdot  10^{-2})$    & \rowcolor{baseline} $8.4_{\pm 0.7}$ & $17.7_{\pm 2.0}$ & $10.3_{\pm 0.6}$ & $17.7_{\pm 2.0}$ & $17.2_{\pm 1.0}$ & $8.4_{\pm 0.7}$ & $\mathbf{5.3_{\pm 1.1}}$ & \rowcolor{our} $17.1_{\pm 1.4}$ & $15.5_{\pm 1.1}$ & $\mathbf{5.3_{\pm 1.1}}$ \\

\textsc{N30} $(\cdot  10^{-2})$     & \rowcolor{baseline} $10.7_{\pm 1.9}$ & $\mathbf{10.2_{\pm 0.1}}$ & $16.1_{\pm 0.1}$ & $\mathbf{10.2_{\pm 0.1}}$ & $\mathbf{10.2_{\pm 0.1}}$ & $10.3_{\pm 0.0}$ & $10.8_{\pm 1.9}$ & \rowcolor{our} $10.3_{\pm 0.1}$ & $10.3_{\pm 0.1}$ & $10.3_{\pm 0.1}$ \\

\textsc{N60} $(\cdot  10^{-2})$     & \rowcolor{baseline} $16.8_{\pm 0.6}$ & $17.1_{\pm 0.5}$ & $18.8_{\pm 0.2}$ & $17.1_{\pm 0.5}$ & $15.0_{\pm 0.1}$ & $17.9_{\pm 0.5}$ & $16.9_{\pm 1.5}$ & \rowcolor{our} $16.1_{\pm 0.6}$ & $15.1_{\pm 0.3}$ & $\mathbf{13.5_{\pm 1.4}}$ \\
\bottomrule
\end{NiceTabular}
}
\end{table*}

\subsection{Forecasting}
\label{sec:forecasting_experiments}
The time series forecasting tasks considered are Lorenz96 \cite{lorenz1996predictability}, Mackey-Glass \cite{jaeger2004esn} and NARMA.

\subsubsection*{Methodology} Similarly to memory-based experiments, we employ the NRMSE as our target metric. Training, validation, and test splits are described individually for each dataset. At training and inference time, we employ a $200$ time step washout to warm up the reservoir. The readout is fed the states at each time step.

\subsubsection*{Lorenz96} The Lorenz96 (Lz) task is to predict the next state of the time series $\mathbf{x}(t)$, expressed as the following $5$-dimensional chaotic system:
\begin{equation}
    \frac{\partial}{\partial t} f_{i}(t) = f_{i-1}(t) ( f_{i+1}(t) - f_{i-2}(t) ) - f_{i}(t) + 8,
    \label{eq:lorenz96}
\end{equation}
for $i = 1, \ldots, 5$. We focus on predicting the 25th (Lz25) and 50th (Lz50) future state of the time series. Thus, the task involves predicting $\mathbf{y}(t) = \mathbf{x}(t + 25)$ and $\mathbf{y}(t) = \mathbf{x}(t + 50)$, respectively. In order to assess the models' performance with a limited number of data points, we generate a time series of length $T = 1200$. The first $400$ time steps are used for training, the next $400$ for validation, and the final $400$ for testing.

\subsubsection*{Mackey-Glass} The Mackey-Glass (MG) task is to predict the next state of the following time series:
\begin{equation}
    \mathbf{x}(t) = \frac{\partial}{\partial t} f(t) = \frac{0.2  f(t - 17)}{1 + f(t - 17)^{10} - 0.1  f(t)}.
    \label{eq:mackey_glass}
\end{equation}
In our experiments, we focus on predicting the 1st and 84th future state of the time series. Thus, the task involves predicting $\mathbf{y}(t) = \mathbf{x}(t + 1)$ (MG) and $\mathbf{y}(t) = \mathbf{x}(t + 84)$ (MG84), respectively. We generate a time series of length $T = 10000$, with the first $5000$ time steps used for training, the next $2500$ for validation, and the final $2500$ for testing.

\subsubsection*{NARMA} Given a one-dimensional input time series $\mathbf{x}(t)$ uniformly distributed in $[0, 0.5]$, the NARMA task is to predict the next state of the following time series:
\begin{equation}
\mathbf{y}(t) = 0.3  \mathbf{y}(t - 1) + 0.01 \mathbf{y}(t - 1) \sum_{i = 1}^{d} \mathbf{y}(t - i) + 1.5  \mathbf{x}(t - d) \mathbf{x}(t - 1) + 0.1.
\label{eq:narma30}
\end{equation}
We consider the NARMA30 (N30) and NARMA60 (N60), with (look-ahead) delay of $d = 30$ and $d = 60$, respectively. We generate a time series of length $T = 10000$, with the first $5000$ time steps used for training, the next $2500$ for validation, and the final $2500$ for testing.

\subsubsection*{Discussion} Table~\ref{tab:forecasting_results} presents the results for the forecasting tasks. Deeper architectures often provide no positive or negligible advantage in tasks with relatively low (look-ahead) delay such as Lz25, MG, and N30. However, DeepResESN shows superior performance in those tasks that require predicting states further into the future, such as Lz50, MG84, and N60. We hypothesize that tasks with lower delays may not offer a sufficiently challenging forecasting problem for additional layers to be relevant. Finally, unlike what was observed in memory-based tasks, there are no substantial differences between configurations. The only exception is MG84, where the identity configuration considerably outperforms the others.

\begin{table*}[t]
\caption{Test set results on classification tasks, averaged over $10$ different random initializations. The \textbf{best result} is in bold.}
\label{tab:classification_results}
\centering
\resizebox{\textwidth}{!}{
\footnotesize
\begin{NiceTabular}{@{}l | c c c c c c c | c c c @{}}
\toprule
$\uparrow$ Classification & \rowcolor{baseline} LeakyESN & ES$^{\text{2}}$N & EuESN & ResESN$_{\mathrm{R}}$ & ResESN$_{\mathrm{C}}$ & ResESN$_{\mathrm{I}}$ & DeepESN & \rowcolor{our} DeepResESN$_{\mathrm{R}}$ & DeepResESN$_{\mathrm{C}}$ & DeepResESN$_{\mathrm{I}}$ \\
\midrule
\textsc{Adiac}     & \rowcolor{baseline} $56.0_{\pm 3.0}$ & $35.5_{\pm 4.1}$ & $63.8_{\pm 1.6}$ & $59.4_{\pm 2.4}$ & $57.3_{\pm 1.5}$ & $61.5_{\pm 1.5}$ & $58.4_{\pm 2.1}$ & \rowcolor{our} $62.3_{\pm 3.1}$ & $57.2_{\pm 3.1}$ & $\mathbf{64.9_{\pm 2.9}}$ \\

\textsc{Blink}      & \rowcolor{baseline} $50.5_{\pm 2.7}$ & $49.5_{\pm 1.6}$ & $\mathbf{87.0_{\pm 3.5}}$ & $61.0_{\pm 4.3}$ & $61.2_{\pm 3.8}$ & $74.7_{\pm 4.8}$ & $71.7_{\pm 2.3}$ & \rowcolor{our} $57.4_{\pm 4.4}$ & $66.3_{\pm 2.2}$ & $80.2_{\pm 5.6}$ \\

\textsc{FordA}     & \rowcolor{baseline} $69.2_{\pm 1.3}$ & $63.5_{\pm 1.4}$ & $66.5_{\pm 1.4}$ & $64.0_{\pm 1.4}$ & $63.9_{\pm 1.6}$ & $69.2_{\pm 1.3}$ & $\mathbf{82.3_{\pm 1.6}}$ & \rowcolor{our} $65.1_{\pm 1.2}$ & $67.4_{\pm 4.4}$ & $\mathbf{82.3_{\pm 1.6}}$ \\

\textsc{FordB}     & \rowcolor{baseline} $60.8_{\pm 0.9}$ & $57.0_{\pm 0.9}$ & $64.7_{\pm 0.8}$ & $56.7_{\pm 1.4}$ & $57.2_{\pm 1.0}$ & $60.8_{\pm 0.9}$ & $\mathbf{66.7_{\pm 1.2}}$ & \rowcolor{our} $55.9_{\pm 0.5}$ & $56.0_{\pm 1.3}$ & $\mathbf{66.7_{\pm 1.2}}$ \\

\textsc{Kepler}    & \rowcolor{baseline} $67.0_{\pm 3.0}$ & $48.8_{\pm 1.5}$ & $69.0_{\pm 2.0}$ & $55.4_{\pm 1.9}$ & $52.8_{\pm 2.2}$ & $67.0_{\pm 3.0}$ & $\mathbf{71.3_{\pm 2.6}}$ & \rowcolor{our} $53.9_{\pm 2.5}$ & $54.7_{\pm 3.4}$ & $\mathbf{71.3_{\pm 2.6}}$ \\

\textsc{Libras}    & \rowcolor{baseline} $76.4_{\pm 1.1}$ & $68.8_{\pm 2.0}$ & $\mathbf{80.6_{\pm 2.2}}$ & $78.4_{\pm 1.8}$ & $75.8_{\pm 1.5}$ & $76.4_{\pm 1.1}$ & $74.4_{\pm 1.9}$ & \rowcolor{our} $77.0_{\pm 1.7}$ & $75.4_{\pm 3.6}$ & $74.4_{\pm 1.9}$ \\

\textsc{Mallat}    & \rowcolor{baseline} $78.7_{\pm 3.6}$ & $30.9_{\pm 1.5}$ & $\mathbf{90.8_{\pm 1.0}}$ & $87.0_{\pm 5.8}$ & $81.8_{\pm 3.7}$ & $78.7_{\pm 3.6}$ & $83.7_{\pm 1.5}$ & \rowcolor{our} $82.5_{\pm 7.0}$ & $81.6_{\pm 5.6}$ & $85.0_{\pm 3.3}$ \\

\textsc{sMNIST}    & \rowcolor{baseline} $70.7_{\pm 4.0}$ & $37.9_{\pm 1.1}$ & $\mathbf{86.9_{\pm 0.3}}$ & $81.6_{\pm 1.0}$ & $77.2_{\pm 5.7}$ & $78.0_{\pm 1.9}$ & $80.1_{\pm 2.0}$ & \rowcolor{our} $79.5_{\pm 2.1}$ & $81.9_{\pm 1.5}$ & $79.3_{\pm 1.7}$ \\

\textsc{psMNIST}   & \rowcolor{baseline} $68.3_{\pm 2.8}$ & $78.7_{\pm 0.2}$ & $83.5_{\pm 0.2}$ & $83.8_{\pm 0.3}$ & $\mathbf{84.0_{\pm 0.3}}$ & $79.1_{\pm 1.4}$ & $78.7_{\pm 0.8}$ & \rowcolor{our} $83.4_{\pm 0.5}$ & $\mathbf{84.0_{\pm 0.2}}$ & $78.7_{\pm 0.6}$ \\

\bottomrule
\end{NiceTabular}
}
\end{table*}

\subsection{Classification}
\label{sec:classification_experiments}
The time series classification tasks considered are a selection of classification problems from the UEA \& UCR repository \cite{bagnall2018uea, dau2019ucr}, plus the MNIST dataset \cite{lecun1998mnist}. To process MNIST as a time series, the images are flattened into one-dimensional vectors, creating the so-called sequential MNIST (sMNIST) task. We also consider the permuted sequential MNIST (psMNIST) task, a slight variation of sMNIST where a random permutation is applied to pixels.

\subsubsection*{Methodology} The validation set is obtained via a $95-5$ stratified split for sMNIST and psMNIST, and via a $70-30$ stratified split otherwise. After model selection, models are retrained on the entire original training set and evaluated on the test set. The readout is fed the states at the last time step.

\subsubsection*{Discussion} Table~\ref{tab:classification_results} presents the results for time series classification tasks. Deeper models generally outperform their shallower counterparts, with notable improvements in Adiac, Blink, FordA, FordB, and Kepler. The only exception is EuESN, which, being conceptually optimized for time series classification, outperforms DeepResESN on Blink, Libras, Mallat, and sMNIST. Among the different configurations, the identity configuration often outperforms both random orthogonal and cyclic orthogonal variants. Specifically, DeepResESN$_{\mathrm{R}}$ and DeepResESN$_{\mathrm{C}}$ show minimal improvements over their shallow counterparts and occasionally yield worse results. In contrast, DeepResESN$_{\mathrm{I}}$ achieves consistent improvements across nearly all datasets. We hypothesize that this superiority stems from the identity matrix's inherent property of preserving exact input information as it passes through multiple nonlinear reservoirs, whereas orthogonal matrices preserve only the input's norm. Additionally, this difference may arise from how each configuration encodes temporal representations of the input, as discussed in Section~\ref{subsec:spectral} and graphically illustrated in Fig.~\ref{fig:spectral}. The (strong) filtering effect applied by the identity configuration may simplify the downstream task by providing the readout with much more diverse output frequencies.

\section{Conclusions}
\label{sec:conclusions}
In this paper, we explored the architectural bias of temporal residual connections in deep untrained Recurrent Neural Networks (RNNs), focusing on the Reservoir Computing (RC) framework. We introduced Deep Residual Echo State Networks (DeepResESNs), a novel class of deep untrained RNNs in which temporal residual connections are integrated within a hierarchy of untrained recurrent layers. We explored various orthogonal configurations for the temporal residual connections, including deterministic fixed-structure variants, such as the cyclic and identity matrices, and randomly generated variants. We analyzed their distinct effects on network dynamics in progressively deeper layers through the lens of spectral frequency analysis, which revealed that each configuration induces a specific filtering behavior on the temporal representation of the input. These diverse behaviors translate into complementary strengths across different tasks, with non-identity orthogonal configurations excelling in memory-based tasks, while the identity configuration proves more effective for classification tasks. Through the lens of linear stability analysis, we formalized necessary and sufficient conditions for obtaining stable and contractive dynamics in DeepResESN. Empirically, our approach consistently outperforms traditional shallow and deep RC across memory-based, forecasting, and classification tasks.

Future work will investigate alternative configurations for the temporal residual connections, and the integration of residual mappings along the spatial dimension.

\section*{Acknowledgments}
This work has been supported by EMERGE, a project funded by the European Innovation Council (prj. code 101070918), and by NEURONE, a project funded by the European Union - Next Generation EU, M4C1 CUP I53D23003600006, under program PRIN 2022 (prj. code 20229JRTZA).

\appendices
\section{Proofs}
\label{sec:proofs}

\subsection{Proof of Theorem~\ref{th:necessary_esp}}
\label{subsec:proof_necessary_esp}
Let us prove \eqref{eq:deepresesn_global_rho}. Recall the global Jacobian of a DeepResESN being a lower triangular block matrix, as defined in \eqref{eq:deepresesnF_jacobian}. Due to such a specific structure, its eigenvalues correspond to the union of the eigenvalues of the matrices along its diagonal:
\begin{equation}
    eig \big( \mathbf{J}_{F, \mathbf{h}} (\mathbf{x}(t), \mathbf{h}(t-1)) \big) = \bigcup_{l=1}^{N_L} eig \big( \mathbf{J}_{F^{(l)}, \mathbf{h}^{(l)}} (\mathbf{x}(t), \mathbf{h}(t-1)) \big),
    \label{jacobian_eigenvalues_union}
\end{equation}
where $eig(\cdot)$ denotes the eigenvalues of its argument.
Thus, the spectral radius of the Jacobian corresponds to the maximum spectral radius among those of the matrices along the diagonal:
\begin{equation}
\rho \big( \mathbf{J}_{F, \mathbf{h}} (\mathbf{x}(t), \mathbf{h}(t-1)) \big) = \max_{l=1, \hdots, N_L} \rho \big( \mathbf{J}_{F^{(l)}, \mathbf{h}^{(l)}} (\mathbf{x}(t), \mathbf{h}(t-1)) \big).
\label{eq:deepresesn_rho_2}
\end{equation}
Consider \eqref{eq:deepresesnFi_jacobian}, the partial derivative of the inter-layer Jacobian for layer $l$ when $i = j = l$ can be rewritten as:
\begin{align}
&\mathbf{J}_{F^{(l)}, \mathbf{h}^{(l)}} (\mathbf{x}(t),  \mathbf{h}(t-1)) \notag \\
&\begin{aligned}
    = \frac{\partial}{\partial \mathbf{h}^{(l)}(t-1)} \big[& \alpha^{(l)} \mathbf{O} \mathbf{h}^{(l)}(t-1) \\
    &\begin{aligned}
        + \beta^{(l)}  \tanh \big( & \mathbf{W}_h^{(l)} \mathbf{h}^{(l)}(t-1) \\
        &+ \mathbf{W}_x^{(l)} F^{(l-1)} ( \mathbf{x}(t), \{\mathbf{h}^{(k)}(t-1)\}_{k=1}^{l-1} ) \\
        &+ \mathbf{b}^{(l)} \big) \big]
    \end{aligned} 
\end{aligned} \notag \\
&\begin{aligned}
=\ &\alpha^{(l)} \mathbf{O} + \beta^{(l)} \left( \begin{array}{ccc}
1 - (\hat{h}_{1}^{(l)}(t))^2 & & \multirow{2}{*}{\text{\huge 0}} \\
& \ddots & \\
\multirow{2}{*}{\vspace{5em} \text{\huge 0}} & & 1 - (\hat{h}_{N_h}^{(l)}(t))^2
\end{array} \right) \mathbf{W}_h^{(l)},
\end{aligned}
\label{eq:diagonal_block_matrices}
\end{align}
where $\hat{h}_{j}^{(l)}(t)$ denotes the j-th element of the post-activation vector $\mathbf{\hat{h}}^{(l)}(t)$, defined as:
\begin{equation}
\begin{aligned}
    \mathbf{\hat{h}}^{(l)}(t) = \tanh \big(& \mathbf{W}_h^{(l)} \mathbf{h}^{(l)}(t-1) \\
    &+ \mathbf{W}_x^{(l)} F^{(l-1)} ( \mathbf{x}(t), \{\mathbf{h}^{(k)}(t-1)\}_{k=1}^{l-1} ) \\
    &+ \mathbf{b}^{(l)} \big).
    \label{eq:post_activation}
\end{aligned}
\end{equation}
Assuming zero input and zero state, with \eqref{eq:diagonal_block_matrices} we derive:
\begin{equation}
\rho(\mathbf{J}_{F^{(l)}, \mathbf{h}^{(l)}} (\mathbf{0}_x, \mathbf{0})) = \rho (\alpha^{(l)}  \mathbf{O} + \beta^{(l)}  \mathbf{W}_h^{(l)} ).
\label{eq:deepresesn_rho_3}
\end{equation}
Plugging \eqref{eq:deepresesn_rho_3} into \eqref{eq:deepresesn_rho_2} we have:
\begin{equation}
    \rho(\mathbf{J}_{F, \mathbf{h}} (\mathbf{0}_x, \mathbf{0})) = \max_{l=1, \hdots, N_L} \rho (\alpha^{(l)}  \mathbf{O} + \beta^{(l)}  \mathbf{W}_h^{(l)} ).
    \label{eq:global_rho}
\end{equation}

Following the necessary condition for the stability of linearized systems around the zero state from \cite{gallicchio2017deepesnesp}, which requires the global spectral radius to be smaller than $1$, the necessary condition for stability around the zero state is:

\begin{equation}
    \rho(\mathbf{J}_{F, \mathbf{h}} (\mathbf{0}_x, \mathbf{0})) = \max_{l=1, \hdots, N_L} \rho ( \alpha^{(l)}  \mathbf{O} + \beta^{(l)} \mathbf{W}_h^{(l)} ) < 1.
    \label{eq:global_rho_stability_cond}
\end{equation}
Following reasoning similar to that presented in \cite{jaeger2001echo, gallicchio2017deepesnesp}, we observe that if the zero state is not a stable fixed point for the state transition function in \eqref{eq:deepresesnF}, then there exists a state $\mathbf{h}_{0}^{'}$ in the neighborhood of the zero state $\mathbf{0}$ such that, given the same null input sequence, the network starting from $\mathbf{h}_{0}^{'}$ will not converge to $\mathbf{0}$. Instead, it will follow a different trajectory compared to a system starting from $\mathbf{0}$. Thus, the ESP condition in \eqref{eq:deepresesnESP} would be violated, and \eqref{eq:deeresesn_necessaryESP} provides a necessary condition for ensuring the ESP of a DeepResESN.

%
\subsection{Proof of Lemma~\ref{lemma:layer_contractivity}}
\label{subsec:proof_layer_contractivity}
The proof can be organized in two cases.

\subsubsection*{case (i)} for $l = 1$, DeepResESN becomes a standard shallow ResESN. We have that $\forall \mathbf{x} \in \mathbb{R}^{N_x},  \forall \mathbf{h}^{(1)}, \mathbf{h}^{'(1)} \in \mathbb{R}^{N_h}$:
\begin{align}
&\lVert F^{(1)} (\mathbf{x}, \mathbf{h}^{(1)}) - F^{(1)} (\mathbf{x}, \mathbf{h}^{'(1)}) \rVert \notag \\
&\begin{aligned}
    &\begin{aligned}
    = \lVert & \alpha^{(1)} \mathbf{O} ( \mathbf{h}^{(1)} - \mathbf{h}^{'(1)} ) \\
        &\begin{aligned}
            + \beta^{(1)} [& \tanh ( \mathbf{W}_h^{(1)} \mathbf{h}^{(1)} + \mathbf{W}_x^{(1)} \mathbf{x} + \mathbf{b}^{(1)} ) \\
            &- \tanh ( \mathbf{W}_h^{(1)} \mathbf{h}^{'(1)} + \mathbf{W}_x^{(1)} \mathbf{x} + \mathbf{b}^{(1)} )] \rVert
        \end{aligned}
    \end{aligned}
\end{aligned} \notag \\
&\leq \alpha^{(1)} \lVert \mathbf{h}^{(1)} - \mathbf{h}^{'(1)}\rVert + \beta^{(1)} \lVert \mathbf{W}_h^{(1)} \rVert  \lVert \mathbf{h}^{(1)} - \mathbf{h}^{'(1)} \rVert \notag \\
&= ( \alpha^{(1)} + \beta^{(1)} \lVert \mathbf{W}_h^{(1)} \rVert ) \lVert \mathbf{h}^{(1)} - \mathbf{h}^{'(1)} \rVert.
\label{eq:lemma1_case_one}
\end{align}
It follows that $C^{(1)} = \alpha^{(1)} + \beta^{(1)} \lVert \mathbf{W}_h^{(1)} \rVert$ is a Lipschitz constant for $F^{(1)}$, which realizes a contraction mapping if $C^{(1)} < 1$.

\subsubsection*{case (ii)} for $l > 1$, the hierarchical structure must be taken into account. Assuming $F^{(l-1)}$ to be a contraction with Lipschitz constant $C^{(l-1)} < 1$, we have that $\forall \mathbf{x} \in 
\mathbb{R}^{N_x},  \forall \{\mathbf{h}^{(k)}\}_{k=1}^{l},\{\mathbf{h}^{'(k)}\}_{k=1}^{l} \in \mathbb{R}^{N_h}$:
\begin{align}
& \lVert F^{(l)} ( \mathbf{x}, \{\mathbf{h}^{(k)}\}_{k=1}^{l} ) - F^{(l)} ( \mathbf{x}, \{\mathbf{h}^{'(k)}\}_{k=1}^{l}) \rVert \notag \\
& \begin{aligned}
=\lVert& \alpha^{(l)} \mathbf{O} ( \mathbf{h}^{(l)} - \mathbf{h}^{'(l)} ) \\
    &\begin{aligned}
    + \beta^{(l)} [& \tanh ( \mathbf{W}_h^{(l)} \mathbf{h}^{(l)} + \mathbf{W}_x^{(l)} F^{(l-1)} ( \mathbf{x}, \{\mathbf{h}^{(k)}\}_{k=1}^{l-1} ) + \mathbf{b}^{(l)} ) \\
        &\begin{aligned}
            - \tanh (& \mathbf{W}_h^{(l)} \mathbf{h}^{'(l)} + \mathbf{W}_x^{(l)} F^{(l - 1)} ( \mathbf{x}, \{\mathbf{h}^{'(k)}\}_{k=1}^{l-1} ) \\
            &+ \mathbf{b}^{(l)} ) ] \rVert
        \end{aligned}
    \end{aligned}
\end{aligned} \notag \\
& \begin{aligned}
\leq \alpha^{(l)} \lVert \mathbf{h}^{(l)} - \mathbf{h}^{'(l)} \rVert + \beta^{(l)} \lVert& \mathbf{W}_h^{(l)} ( \mathbf{h}^{(l)} - \mathbf{h}^{'(l)} ) \\
    &\begin{aligned}
    + \mathbf{W}_x^{(l)} [& F^{(l-1)} ( \mathbf{x}, \{\mathbf{h}^{(k)}\}_{k=1}^{l-1} ) \\
    &- F^{(l-1)} ( \mathbf{x}, \{\mathbf{h}^{'(k)}\}_{k=1}^{l-1} ) ] \rVert
    \end{aligned}
\end{aligned} \notag \\
& \begin{aligned}
\leq \alpha^{(l)} \lVert \mathbf{h}^{(l)} - \mathbf{h}^{'(l)} \rVert + \beta^{(l)} (& \lVert \mathbf{W}_h^{(l)} \rVert \lVert \mathbf{h}^{(l)} - \mathbf{h}^{'(l)} \rVert \\
&\begin{aligned}
    + C^{(l-1)} \lVert \mathbf{W}_x^{(l)} \rVert \lVert &\{\mathbf{h}^{(k)}\}_{k=1}^{l-1} \\
    &- \{\mathbf{h}^{'(k)}\}_{k=1}^{l-1} \rVert )
\end{aligned}
\end{aligned} \notag \\
& \begin{aligned}
\leq & \alpha^{(l)} \lVert \{\mathbf{h}^{(k)}\}_{k=1}^{l} - \{\mathbf{h}^{'(k)}\}_{k=1}^{l} \rVert \notag \\
&\begin{aligned}
    + \beta^{(l)} (& \lVert \mathbf{W}_h^{(l)} \rVert \lVert \{\mathbf{h}^{(k)}\}_{k=1}^{l} - \{\mathbf{h}^{'(k)}\}_{k=1}^{l} \rVert \\
    &+ C^{(l-1)} \lVert \mathbf{W}_x^{(l)} \rVert \lVert \{\mathbf{h}^{(k)}\}_{k=1}^{l} - \{\mathbf{h}^{'(k)}\}_{k=1}^{l} \rVert )
\end{aligned}
\end{aligned} \notag \\
& \begin{aligned}
    = [ \alpha^{(l)} + \beta^{(l)} ( \lVert \mathbf{W}_h^{(l)} \rVert + C^{(l-1)}\lVert \mathbf{W}_x^{(l)} \rVert ) ] \, \lVert &\{\mathbf{h}^{(k)}\}_{k=1}^{l} \\
    &- \{\mathbf{h}^{'(k)}\}_{k=1}^{l} \rVert.
\end{aligned}
\label{eq:lemma1_case_two}
\end{align}
It follows that $C^{(l)} = \alpha^{(l)} + \beta^{(l)} ( \lVert \mathbf{W}_h^{(l)} \rVert + C^{(l-1)} \lVert \mathbf{W}_x^{(l)} \rVert )$ is a Lipschitz constant for $F^{(l)}$. Thus, $F^{(l)}$ realizes a contraction mapping for $C^{(l)} < 1$.

\subsection{Proof of Theorem~\ref{th:sufficient_esp}}
\label{subsec:proof_sufficient_esp}
Let us prove that \eqref{eq:contractivity_global} ensures global contractive dynamics. We have that $\forall \mathbf{x} \in \mathbb{R}^{N_x},  \forall \{\mathbf{h}^{(k)}\}_{k=1}^{N_L}, \{\mathbf{h}^{'(k)}\}_{k=1}^{N_L} \in \mathbb{R}^{N_h}$:
\begin{align}
& \lVert F ( \mathbf{x}, \{\mathbf{h}^{(k)}\}_{k=1}^{N_L} ) 
  - F ( \mathbf{x}, \{\mathbf{h}^{'(k)}\}_{k=1}^{N_L} ) \rVert \notag \\
&\begin{aligned} 
    = \lVert &[ F^{(1)} ( \mathbf{x}, \{\mathbf{h}^{(k)}\}_{k=1}^{1} ), \ldots, F^{(N_L)} ( \mathbf{x}, \{\mathbf{h}^{(k)}\}_{k=1}^{N_L} ) ] \\
    &- [ F^{(1)} ( \mathbf{x}, \{\mathbf{h}^{'(k)}\}_{k=1}^{1} ), \ldots, F^{(N_L)} ( \mathbf{x}, \{\mathbf{h}^{'(k)}\}_{k=1}^{N_L} ) ] \rVert
\end{aligned} \notag \\
&= \max_{l=1, \hdots, N_L} [ \lVert F^{(l)} ( \mathbf{x}, \{\mathbf{h}^{(k)}\}_{k=1}^{l} ) -  F^{(l)} ( \mathbf{x}, \{\mathbf{h}^{'(k)}\}_{k=1}^{l} ) \rVert ] \notag \\
&\leq \max_{l=1, \hdots, N_L} ( C^{(l)} \lVert \{\mathbf{h}^{(k)}\}_{k=1}^{l} - \{\mathbf{h}^{'(k)}\}_{k=1}^{l} \rVert ) \notag \\
&\leq \max_{l=1, \hdots, N_L} ( C^{(l)} )  \lVert \{\mathbf{h}^{(k)}\}_{k=1}^{N_L} - \{\mathbf{h}^{'(k)}\}_{k=1}^{N_L} \rVert.
\label{eq:contractivity_global_cond}
\end{align}
Thus $C = \max_{l=1, \hdots, N_L} ( C^{(l)} )$ is a Lipschitz constant for $F$, and $F$ is a contraction mapping if $C < 1$. Consider the iterated version of the state transition function $\hat{F}$ defined in \eqref{eq:deepresesnF_iterated}. Assume any input sequence of length $T$, denoted as $\mathbf{s}_{T} = \{\mathbf{x}_k\}_{k=1}^{T}$. Then, we have that $\forall \{\mathbf{h}^{(k)}\}_{k=1}^{N_L},  \{\mathbf{h}^{'(k)}\}_{k=1}^{N_L} \in \mathbb{R}^{N_L \times N_h}$:
\begin{align}
&\lVert \hat{F}(\mathbf{s}_{T}, \mathbf{h}) - \hat{F}(\mathbf{s}_{T}, \mathbf{h}^{'}) \rVert \notag \\
&= \lVert \hat{F} ( \{\mathbf{x}_{k}\}_{k=1}^{T}, \mathbf{h} ) - \hat{F} ( \{\mathbf{x}_{k}\}_{k=1}^{T}, \mathbf{h}^{'} ) \rVert \notag \\
&= \lVert F [ \mathbf{x}_{T}, \hat{F} ( \{\mathbf{x}_k\}_{k=1}^{T-1}, \mathbf{h} ) ] - F [ \mathbf{x}_{T}, \hat{F} ( \{\mathbf{x}_k\}_{k=1}^{T-1}, \mathbf{h}^{'} ) ] \rVert \notag \\
&\leq C  \lVert \hat{F} ( \{\mathbf{x}_k\}_{k=1}^{T-1}, \mathbf{h} ) - \hat{F} ( \{\mathbf{x}_k\}_{k=1}^{T-1}, \mathbf{h}^{'} ) \rVert \notag \\
&= C \lVert F [ \mathbf{x}_{T-1}, \hat{F} ( \{\mathbf{x}_k\}_{k=1}^{T - 2}, \mathbf{h} ) ] - F [ \mathbf{x}_{T-1}, \hat{F} ( \{\mathbf{x}_k\}_{k=1}^{T - 2}, \mathbf{h}^{'} ) ] \rVert \notag \\
&\leq C^{2}  \lVert \hat{F} ( \{\mathbf{x}_k\}_{k=1}^{T-2}, \mathbf{h} ) - \hat{F} ( \{\mathbf{x}_k\}_{k=1}^{T-2}, \mathbf{h}^{'} ) \rVert \notag \\
& \hdots \notag \\
&\leq C^{T-1}  \lVert \hat{F}(\mathbf{x}_1, \mathbf{h}) - \hat{F}(\mathbf{x}_1, \mathbf{h}^{'}) \rVert \notag \\
&= C^{T-1}  \lVert F \big( \mathbf{x}_1, \hat{F}([ \; ], \mathbf{h}) \big) - F \big( \mathbf{x}_1, \hat{F} ([ \; ], \mathbf{h}^{'}) \big) \rVert \notag \\
&= C^{T-1}  \lVert F(\mathbf{x}_1, \mathbf{h}) - F(\mathbf{x}_1, \mathbf{h}^{'}) \rVert \notag \\
&\leq C^{T}  \lVert \mathbf{h} - \mathbf{h}^{'} \rVert \notag \\
&= C^{T}  \max_{l=1, \hdots, N_L} ( \lVert \mathbf{h}^{(l)} - \mathbf{h}^{'(l)} \rVert ) \notag \\
&\leq C^{T}  D.
\label{eq:proof_sufficient_esp}
\end{align}
where $D$ is a diameter that bounds the reservoir global state space. Observe that $\lim_{T \rightarrow \infty} C^{T}  D = 0$, and that $\lVert \hat{F}(\mathbf{s}_{T}, \mathbf{h}) - \hat{F}(\mathbf{s}_{T}, \mathbf{h}^{'}) \rVert$ is upper-bounded by $C^{T} D$ due to the last inequality in \eqref{eq:proof_sufficient_esp}. Recalling \eqref{eq:deepresesnESP}, the ESP is satisfied for all inputs.

\bibliographystyle{IEEEtran}
\bibliography{bibliography}

@article{lecun2015deep,
  title={Deep learning},
  author={LeCun, Yann and Bengio, Yoshua and Hinton, Geoffrey},
  journal={nature},
  volume={521},
  number={7553},
  pages={436--444},
  year={2015},
  publisher={Nature Publishing Group UK London}
}

@article{vaswani2017attention,
  title={Attention is all you need},
  author={Vaswani, Ashish and Shazeer, Noam and Parmar, Niki and Uszkoreit, Jakob and Jones, Llion and Gomez, Aidan N and Kaiser, {\L}ukasz and Polosukhin, Illia},
  journal={Advances in neural information processing systems},
  volume={30},
  year={2017}
}

@article{simonyan2015deep,
  title={Very deep convolutional networks for large-scale image recognition},
  author={Simonyan, Karen and Zisserman, Andrew},
  journal={International Conference on Learning Representations},
  year={2015}
}

@article{bengio1994gradient,
  title={Learning long-term dependencies with gradient descent is difficult},
  author={Bengio, Yoshua and Simard, Patrice and Frasconi, Paolo},
  journal={IEEE transactions on neural networks},
  volume={5},
  number={2},
  pages={157--166},
  year={1994},
  publisher={IEEE}
}

@inproceedings{glorot2010understanding,
  title={Understanding the difficulty of training deep feedforward neural networks},
  author={Glorot, Xavier and Bengio, Yoshua},
  booktitle={Proceedings of the thirteenth international conference on artificial intelligence and statistics},
  pages={249--256},
  year={2010},
  organization={JMLR Workshop and Conference Proceedings}
}

@inproceedings{he2016resnet,
  title={Deep residual learning for image recognition},
  author={He, Kaiming and Zhang, Xiangyu and Ren, Shaoqing and Sun, Jian},
  booktitle={Proceedings of the IEEE conference on computer vision and pattern recognition},
  pages={770--778},
  year={2016}
}

@article{shi2021novel,
  title={Novel discrete-time recurrent neural network for robot manipulator: A direct discretization technical route},
  author={Shi, Yang and Zhao, Wenhan and Li, Shuai and Li, Bin and Sun, Xiaobing},
  journal={IEEE Transactions on Neural Networks and Learning Systems},
  volume={34},
  number={6},
  pages={2781--2790},
  year={2021},
  publisher={IEEE}
}

@article{verstraeten2007reservoir,
  title={An experimental unification of reservoir computing methods},
  author={Verstraeten, David and Schrauwen, Benjamin and d’Haene, Michiel and Stroobandt, Dirk},
  journal={Neural networks},
  volume={20},
  number={3},
  pages={391--403},
  year={2007},
  publisher={Elsevier}
}

@article{yildiz2012re,
  title={Re-visiting the echo state property},
  author={Yildiz, Izzet B and Jaeger, Herbert and Kiebel, Stefan J},
  journal={Neural networks},
  volume={35},
  pages={1--9},
  year={2012},
  publisher={Elsevier}
}

@article{gallicchio2017deepesn,
  title={Deep reservoir computing: A critical experimental analysis},
  author={Gallicchio, Claudio and Micheli, Alessio and Pedrelli, Luca},
  journal={Neurocomputing},
  volume={268},
  pages={87--99},
  year={2017},
  publisher={Elsevier}
}

@article{chang2017dilated,
  title={Dilated recurrent neural networks},
  author={Chang, Shiyu and Zhang, Yang and Han, Wei and Yu, Mo and Guo, Xiaoxiao and Tan, Wei and Cui, Xiaodong and Witbrock, Michael and Hasegawa-Johnson, Mark A and Huang, Thomas S},
  journal={Advances in neural information processing systems},
  volume={30},
  year={2017}
}

@article{ceni2025random,
  title={Random orthogonal additive filters: a solution to the vanishing/exploding gradient of deep neural networks},
  author={Ceni, Andrea},
  journal={IEEE Transactions on Neural Networks and Learning Systems},
  volume={36},
  year={2025},
  number={6},
  pages={10794-10807},
}

@article{ceni2024residual,
  title={Residual Echo State Networks: Residual recurrent neural networks with stable dynamics and fast learning},
  author={Ceni, Andrea and Gallicchio, Claudio},
  journal={Neurocomputing},
  volume={597},
  pages={127966},
  year={2024},
  publisher={Elsevier}
}

@inproceedings{pinna2025residual,
  title={Residual reservoir memory networks},
  author={Pinna, Matteo and Ceni, Andrea and Gallicchio, Claudio},
  booktitle={International Joint Conference on Neural Networks (IJCNN)},
  pages={1--7},
  year={2025},
  organization={IEEE}
}

@article{gallicchio2024euler,
  title={Euler state networks: Non-dissipative reservoir computing},
  author={Gallicchio, Claudio},
  journal={Neurocomputing},
  volume={579},
  pages={127411},
  year={2024},
  publisher={Elsevier}
}

@article{jaeger2007leakyesn,
  title={Optimization and applications of echo state networks with leaky-integrator neurons},
  author={Jaeger, Herbert and Luko{\v{s}}evi{\v{c}}ius, Mantas and Popovici, Dan and Siewert, Udo},
  journal={Neural networks},
  volume={20},
  number={3},
  pages={335--352},
  year={2007},
  publisher={Elsevier}
}

@article{ceni2024edge,
  title={Edge of stability echo state network},
  author={Ceni, Andrea and Gallicchio, Claudio},
  journal={IEEE Transactions on Neural Networks and Learning Systems},
  volume={36},
  number={4},
  pages={7555--7564},
  year={2024},
  publisher={IEEE}
}

@article{rodan2010minimum,
  title={Minimum complexity echo state network},
  author={Rodan, Ali and Tino, Peter},
  journal={IEEE transactions on neural networks},
  volume={22},
  number={1},
  pages={131--144},
  year={2010},
  publisher={IEEE}
}

@article{tino2020dynamical,
  title={Dynamical systems as temporal feature spaces},
  author={Tino, Peter},
  journal={Journal of Machine Learning Research},
  volume={21},
  number={44},
  pages={1--42},
  year={2020}
}

@article{koryakin2012balanced,
  title={Balanced echo state networks},
  author={Koryakin, Danil and Lohmann, Johannes and Butz, Martin V},
  journal={Neural Networks},
  volume={36},
  pages={35--45},
  year={2012},
  publisher={Elsevier}
}

@article{otte2016optimizing,
  title={Optimizing recurrent reservoirs with neuro-evolution},
  author={Otte, Sebastian and Butz, Martin V and Koryakin, Danil and Becker, Fabian and Liwicki, Marcus and Zell, Andreas},
  journal={Neurocomputing},
  volume={192},
  pages={128--138},
  year={2016},
  publisher={Elsevier}
}

@inproceedings{frigo98fft,
  title={An adaptive software architecture for the FFT},
  author={Frigo, Matteo and Johnson, S},
  booktitle={Proc. ICASSP},
  volume={98},
  pages={1381},
  year={1998}
}

@article{gallicchio2017deepesnesp,
  title={Echo state property of deep reservoir computing networks},
  author={Gallicchio, Claudio and Micheli, Alessio},
  journal={Cognitive Computation},
  volume={9},
  pages={337--350},
  year={2017},
  publisher={Springer}
}

@article{jaeger2001echo,
  title={The “echo state” approach to analysing and training recurrent neural networks-with an erratum note},
  author={Jaeger, Herbert},
  journal={Bonn, Germany: German National Research Center for Information Technology GMD Technical Report},
  volume={148},
  number={34},
  pages={13},
  year={2001},
  publisher={Bonn}
}

@article{bagnall2018uea,
  title={The UEA multivariate time series classification archive},
  author={Bagnall, Anthony and Dau, Hoang Anh and Lines, Jason and Flynn, Michael and Large, James and Bostrom, Aaron and Southam, Paul and Keogh, Eamonn},
  journal={arXiv preprint arXiv:1811.00075},
  year={2018}
}

@article{dau2019ucr,
  title={The UCR time series archive},
  author={Dau, Hoang Anh and Bagnall, Anthony and Kamgar, Kaveh and Yeh, Chin-Chia Michael and Zhu, Yan and Gharghabi, Shaghayegh and Ratanamahatana, Chotirat Ann and Keogh, Eamonn},
  journal={IEEE/CAA Journal of Automatica Sinica},
  volume={6},
  number={6},
  pages={1293--1305},
  year={2019},
  publisher={IEEE}
}

@inproceedings{verstraeten2010ctxor,
  title={Memory versus non-linearity in reservoirs},
  author={Verstraeten, David and Dambre, Joni and Dutoit, Xavier and Schrauwen, Benjamin},
  booktitle={International joint conference on neural networks (IJCNN)},
  pages={1--8},
  year={2010},
  organization={IEEE}
}

@article{inubushi2017sinmem,
  title={Reservoir computing beyond memory-nonlinearity trade-off},
  author={Inubushi, Masanobu and Yoshimura, Kazuyuki},
  journal={Scientific reports},
  volume={7},
  number={1},
  pages={10199},
  year={2017},
  publisher={Nature Publishing Group UK London}
}

@article{jaeger2002memory,
  title={Short term memory in echo state networks},
  author={Jaeger, Herbert},
  year={German Nat. Res. Center Inform. Technol., St. Augustin, Germany, Tech. Rep. GMD 152, 2002},
}

@inproceedings{lorenz1996predictability,
  title={Predictability: A problem partly solved},
  author={Lorenz, Edward N},
  booktitle={Proc. Seminar on predictability},
  volume={1},
  number={1},
  year={1996},
  organization={Reading}
}

@article{jaeger2004esn,
  title={Harnessing nonlinearity: Predicting chaotic systems and saving energy in wireless communication},
  author={Jaeger, Herbert and Haas, Harald},
  journal={science},
  volume={304},
  number={5667},
  pages={78--80},
  year={2004},
  publisher={American Association for the Advancement of Science}
}

@article{lecun1998mnist,
  title={The MNIST database of handwritten digits},
  author={LeCun, Yann},
  journal={http://yann. lecun. com/exdb/mnist/},
  year={1998}
}

@article{demvsar2006statistical,
  title={Statistical comparisons of classifiers over multiple data sets},
  author={Dem{\v{s}}ar, Janez},
  journal={Journal of Machine learning research},
  volume={7},
  number={Jan},
  pages={1--30},
  year={2006}
}

\end{document}